\documentclass[11pt]{murtylab}
\pdfoutput=1
\usepackage{graphicx}
\usepackage{cleveref}
\usepackage{longtable}
\usepackage{tabularx}
\usepackage{bbding}
\usepackage{adjustbox}
\usepackage{array}
\usepackage{multirow}
\usepackage{makecell}

\usepackage{subcaption}
\usepackage{listings}
\usepackage{xcolor}
\usepackage{amsfonts}

\lstdefinestyle{python}{
    language=Python,
    basicstyle=\ttfamily\small,
    keywordstyle=\bfseries\color{blue},
    stringstyle=\color{green!50!black},
    commentstyle=\color{red!50!black},
    emphstyle=\bfseries\color{violet},
    breaklines=true,
    showstringspaces=false,
    frame=single,
    numberstyle=\tiny\color[gray], 
    xleftmargin=15pt,
}

\title{\huge TopoNets: High performing vision and language models with brain-like topography}

\author[1,2]{\Large Mayukh Deb}
\author[3]{\Large Mainak Deb}
\author[1,2]{\Large N. Apurva Ratan Murty}

\usepackage{xspace}

\makeatletter
\DeclareRobustCommand\onedot{\futurelet\@let@token\@onedot}
\def\@onedot{\ifx\@let@token.\else.\null\fi\xspace}

\makeatother

\affiliation[1]{Cognition and Brain Science, School of Psychology, Georgia Tech}

\affiliation[2]{Computational Cognition, Georgia Tech}
\affiliation[3]{Independent contributor}

\abstract{ Neurons in the brain are organized such that nearby cells tend to share similar functions. AI models lack this organization, and past efforts to introduce topography have often led to trade-offs between topography and task performance. In this work, we present \textit{TopoLoss}, a new loss function that promotes spatially organized topographic representations in AI models without significantly sacrificing task performance. TopoLoss is highly adaptable and can be seamlessly integrated into the training of leading model architectures. We validate our method on both vision (ResNet-18, ResNet-50, ViT) and language models (GPT-Neo-125M, NanoGPT), collectively \textit{TopoNets}. TopoNets are the highest performing supervised topographic models to date, exhibiting brain-like properties such as localized feature processing, lower dimensionality, and increased efficiency. TopoNets also predict responses in the brain and replicate the key topographic signatures observed in the brain’s visual and language cortices. Together this work establishes a robust and generalizable framework for integrating topography into leading model architectures, advancing the development of high performing models that more closely emulate the computational strategies of the human brain.}

\date{December 7th, 2024}
\correspondence{First/Last Author at \texttt{\href{mailto:mayukh@gatech.edu,ratan@gatech.edu}{[mayukh/ratan]@gatech.edu}}}
\metadata[Project Website]{\url{https://toponets.github.io}}
\metadata[ICLR 2025 Review Page]{\url{https://openreview.net/forum?id=THqWPzL00e}} 
\begin{document}

\maketitle

\section{Introduction}

Neurons in the brain are not tossed around haphazardly. They're spatially organized such that nearby cells perform similar functions \cite{barlow1986have, rakic1988specification,eickhoff2018topographic,krubitzer2009search}. \textit{Topographic organization} is a core feature of brains \cite{geschwind2013cortical, arcaro2024whole}. In visual cortex this organization is evident in micro-scale pinwheel patterns for orientation selectivity\cite{maldonado1997orientation, bonhoeffer1991iso}, in macro-scale category-selective regions for faces \citep{kanwisher2002fusiform, kanwisher2006fusiform}, bodies \citep{downing2001cortical}, scenes \citep{epstein1998cortical} etc., and in large-scale organizational biases for real-world shape, size, curvature, and animacy \cite{konkle2013tripartite, konkle2011canonical}. Beyond vision in the language cortex, recent studies have also identified clustering of neurons with distinct temporal integration windows \cite{hasson2008hierarchy, lerner2011topographic, regev2024neural}. Unlike the brain, most artificial neural network (ANN) models lack any kind of systematic organization of units. In this work we introduce \textit{TopoLoss}, a new brain-inspired inductive bias that can be easily integrated into the training of most current ANN architectures (both convolutional networks and transformers). The resulting models, \textit{TopoNets}, exhibit brain-like topography characterized by localized, low-dimensional, and efficient representations (Figure 1).

\begin{figure}
    \centering
    \includegraphics[width=0.880\linewidth]{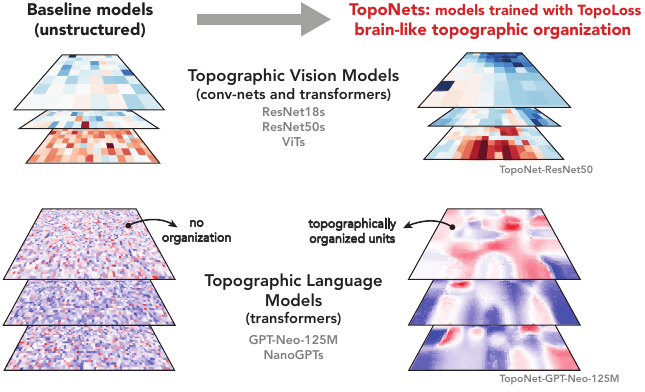}
    \caption{\textbf{Towards high-performing topographic vision and language models (TopoNets).} Schematic shows transformation from unstructured baseline models (left) to organized topographic representations (right) for vision (top) and language models (right). The stacked maps are 3 representative layers (early, mid and late) of the model.}
    \setlength{\belowcaptionskip}{0pt} 
\end{figure}
Inducing topography into artificial neural networks (ANNs) has proven to be challenging and two major strategies have emerged. The first, \textit{post-hoc topography}, involves re-organizing units in pretrained models using methods like self-organizing maps \cite{doshi2023,zhang2021principles,kohonen1997exploration}. The resulting models exhibit topographic signatures, but the underlying representations remain unchanged from the original model. Consequently, the functional advantages of topography, such as reduced dimensionality and increased efficiency, are not realized.  The second strategy, \textit{jointly-optimized topography}, incorporates an additional topographic loss during model training. These models induce topography by either (1) explicitly matching the brain's spatial correlation structure  \cite{lee2020topographic, margalit2024unifying}, (2) imposing distance-dependent constraints \cite{blauch2022connectivity,qian2024local}, or (3) encouraging information redundancy \cite{keller2021modeling}. These approaches suffer a significant tradeoff: the ability of the model to learn task-relevant representations is often compromised. They perform poorly on engineering metrics (like performance on ImageNet) and/or show diminished capacity to predict brain data. Also most prior work in this space has focused exclusively on vision models and the one attempt\cite{binhuraib2024topoformer} at imparting topography to language models focused on self-attention maps of BERT, which resulted in only modest topographic organization in the output space. To summarize, there is currently no unified strategy to apply topography across common ANN architectures (e.g., convolutional nets and transformers) and domains (vision and language) that can deliver \textit{high-performing} models together with the \textit{functional benefits of topography}.

Here we set out to design a general inductive model bias to recapitulate signatures of brain-like topography and its computational benefits without sacrificing model accuracy. To achieve this, it was important to understand \textit{why} and \textit{how} topography arises in the first place. Theoretical work in neuroscience suggests that one of the primary evolutionary pressures on the brain is metabolic efficiency, both in terms of minimizing wiring length of neurons \cite{KAAS1997107} and in managing the vast network of potential neural connections \cite{katz1996synaptic,chklovskii2002wiring, chklovskii2004maps}. The brain addresses this challenge through synaptic pruning. Early in development, the brain forms an excess of synaptic connections, which are then systematically reduced over time based on activity-dependent mechanisms \cite{KAAS1997107,faust2021mechanisms,riccomagno2015sculpting,schulz2005mirror}. This pruning process retains only the most necessary connections, optimizing for the efficiency of the neural network. Our topographic loss incorporates ideas about synaptic pruning into its design.

Our study makes the following contributions: A) We introduce \textit{TopoLoss}, a new inductive bias that generalizes across model architectures (convolutional networks and transformers) and domains (vision and language).
(B) We show that TopoNets, our suite of supervised topographic models, outperform previous topographic models on ImageNet performance and predictions on brain data (as on BrainScore) while maintaining similar levels of topography.
(C) TopoNets provide clear evidence resolving theoretical claims about the role of topography in creating low-dimensional feature representations.
(D) We show that TopoNets demonstrate improved model efficiency.
(E) TopoNets replicate topographic signatures observed in the visual and language cortices in the brain.

\section{Methods}

\subsection{Defining the Cortical Sheet}

Our first task was to define a 2D sheet (See Appendix A.1 for a discussion on topography on 2D versus 3D spaces) where we could apply our topographic loss (\textit{TopoLoss}). To demonstrate that TopoLoss generalizes across different domains, we applied it to both language and vision models. For \textit{language}, we trained GPT-Neo-125M models \cite{gpt-neo} on the Wikipedia Dataset \cite{wikidump} and NanoGPT models  \cite{Karpathy2022} on 10 billion tokens from FineWeb-Edu \cite{lozhkov2024fineweb-edu}. For \textit{vision}, we trained topographic ResNet-18 to allow comparisons with previous topographic models, and ResNet-50 \cite{resnet18} and ViT-b32  \cite{dosovitskiy2020image} models to further evaluate generalization of TopoLoss to larger models and architectures. All vision models were trained on a supervised 1000-way classification task on ImageNet \cite{imagenet}. Together these language and vision models allowed us to robustly evaluate TopoLoss across varied model architectures and domains.

\textbf{Cortical sheet in Transformers (language models and ViTs)}:  For a linear layer with $i$ input units and $o$ output units, we reshape it's weight matrix \( \mathbf{W} \in \mathbb{R}^{o \times i} \) to a cortical sheet  \( \mathbf{C} \in \mathbb{R}^{h \times w \times d} \). In this setup, the  area of the sheet $(h \times w)$ corresponds to the number of output units ($o$) and the depth ($d$) corresponds to the number of input units $(i)$. To maximize the number of neighbors for each element in the cortical sheet, we chose $h$ and $w$ to be as close to each other as possible, thereby minimizing the perimeter.  Each``element'' in this cortical sheet now represents the weights associated with a single output unit (or ``neuron'') in the original linear layer.

\textbf{Cortical sheet in Convolutional Models}: For a convolutional layer with $c_{\textnormal{input}}$ input channels and $c_{\textnormal{output}}$ output channels, and a kernel-size of $k \times k$,  we project its weight tensor \( \mathbf{W} \in \mathbb{R}^{c_{\textnormal{output}}
\times c_{\textnormal{input}}
\times k \times k} \) onto a cortical sheet \( \mathbf{C} \in \mathbb{R}^{h \times w \times d} \), where the area corresponds to the number of output channels ($h \times w$) and the depth is defined as $d = c_{\textnormal{input}} \times k \times k$.  As in previous work \cite{qian2024local}, we arranged the model units (convolutional kernels) on a 2D cortical sheet. A more detailed explanation is provided in the appendix A.2. 

\subsection{Inducing topography (TopoLoss)}
The second step introduces the TopoLoss to the reshaped cortical sheet, promoting topographic organization. We achieve this by maximizing the cosine similarity between the original cortical sheet and its blurred version. This suppresses the high-frequency noise, leaving behind only the most important and meaningful information. This idea was motivated by synaptic pruning in the brain, which eliminates noisy (high frequency) neural connections, refining the biological network's structure (although note that we do not explicitly \textit{remove} any weights here). The blurring of a 2D signal \( X \in \mathbb{R}^{h \times w} \) can be defined using a downsampling function $f_{down}$  and upsampling function $f_{up}$ a  as follows:
\begin{equation}
\textnormal{Blur}(X, \phi_{h}, \phi_{w}) = f_{\textnormal{up}}\left(f_{\textnormal{down}}\left(X, \frac{h}{\phi_h}, \frac{w}{\phi_w}\right), h, w\right)
\end{equation}
Here $\phi_h$ and $\phi_w$ are the downsampling factors along height and width dimensions (both set to 3).  To encourage smoothness in the cortical sheet $C^{h \times w \times d}$  we maximize the cosine similarity between $C$ and its blurred version $C'$ across cortical sheet layers maps. This process smoothens the representations and encourages topographic organization. The TopoLoss is defined as:
\[
\mathcal{L}_{\text{topo}} = -\frac{1}{N} \sum_{i=1}^{N} \frac{C_i \cdot C'_i}{\|C_i\| \|C'_i\|}
\]
This TopoLoss is integrated with the original training loss $\mathcal{L}_{\text{training}}$ as follows:
\[
\mathcal{L}_{\text{total}}  = \mathcal{L}_{\text{training}} + \tau(\mathcal{L}_{\text{topo}})
\]
Here \(\tau\) is a scaling factor that controls the strength of the topographic effect: higher values encourage stronger topographic organization in the model.

\textbf{Vision Models}:  We applied TopoLoss to every convolutional layer in the residual blocks (as \cite{qian2024local}). All vision models were trained on a supervised 1000-way classification task using ImageNet. 

\textbf{ResNet-18} We trained 8 distinct ResNet-18  \cite{resnet18} models from scratch on the ImageNet \cite{imagenet} dataset across various topographic configurations: one baseline model (no topography), six TopoNets with different topographic scaling factors: $\tau = 0.5, 1, 5,10,20,50$. Models were trained using the \texttt{ffcv} \cite{ffcv} training recipe.  \texttt{ffcv} (Fast Forward Computer Vision) significantly accelerates model training by replacing traditional data loaders with an efficient binary format and leveraging multiprocessing and GPU-accelerated data augmentation to optimize data pipelines.

\textbf{ResNet-50:}  We selected ResNet-18 to compare the performance of TopoNets with previous topographic approaches like TDANNs and LLCNN. However it has been demonstrated that ResNet-50 offers a richer visual representational basis for predicting brain responses (see: \cite{ratan2021computational}).
Hence we trained 3 additional ResNet-50  \cite{resnet18} models from scratch on ImageNet \cite{imagenet}:  one baseline model (no topography), two TopoNets with different topographic scaling factors ($\tau = 1, 30$).

\textbf{ViT-b-32} To demonstrate further generalizability beyond convolutional architectures for vision, we trained a Vision Transformer \cite{dosovitskiy2020image}
 on the ImageNet dataset. We followed the recipe provided by TorchVision \cite{torchvision2016} and applied TopoLoss with $\tau = 10$ on the last MLP module i.e the \texttt{mlp.3} module in each transformer block.

 \textbf{Language Models}: We trained two Autoregressive Language models with topoloss applied on the first layer of the feed-forward module in all of the transformer blocks.
 
\textbf{GPT-Neo-125M:}  We trained 5 GPT-Neo-125M \cite{gpt-neo} models on the wikipedia dataset \cite{wikidump} with different scales of the topographic loss (baseline and $\tau$=1, 5, 10 and 50 respectively) .  We applied TopoLoss in the c\_fc layer of GPT-Neo. This choice was based on  prior work by \cite{geva2020transformer, geva2022transformer} that has suggested that the feed-forward modules in GPTs act as key-value memory modules storing world knowledge. The c\_fc modules encode the persistent representations (in contrast to transient representations in the attention matrix) making it the theoretically grounded target for inducing topography.

\textbf{NanoGPT-125M} We trained 4 NanoGPT \cite{Karpathy2022} models on 10 Billion tokens sampled randomly from the FineWeb-Edu dataset \cite{lozhkov2024fineweb-edu} with different scales of the topgraphic loss (baseline and $\tau$ = 0.5, 1, 50 respectively). TopoLoss was applied to the  c\_fc modules in each block (as explained above).

\subsection{Other Metrics}

\textbf{Effective Dimensionality: } Effective dimensionality was measured as described previously in \cite{margalit2024unifying, del2021effective, elmoznino2024high}. 
\[
\text{Effective Dimensionality} = \frac{\left( \sum_{i=1}^{n} \lambda_i \right)^2}{\sum_{i=1}^{n} \lambda_i^2}
\]
$\lambda_i$ indicates the eigenvalues and $n$ the number of eigenvalues. This metric measures the \textit{spread} of the  eigenspectrum. For ResNets, we followed the procedure outlined in \cite{margalit2024unifying}. We chose 20,000 images from the ImageNet validation set calculated the effective dimensionality of the features for all the convolutional layers. For language models, we chose 8192 samples from the openwebtext dataset and measured dimensionality of the representations from the topographic (\texttt{c\_fc}) layers.

\textbf{Smoothness} 
We measured topography, using smoothness score (as before, \cite{margalit2024unifying}).  Smoothness was defined as difference between the highest and lowest correlation values from pairwise correlation versus distance plots.

\textbf{L1 unstructured pruning:} We impose sparsity by pruning a percentage of the smallest-magnitude weights. Specifically, we sort the weights in ascending order of their absolute magnitude and set the smallest $n\%$ of them to zero. To reduce the number of weights by a factor of $n$, we prune $(100 - \frac{100}{n})\%$ of the smallest weights.

\textbf{Downsampling: } We downsample the topographic layers by first projecting them into the cortical space and then performing a downsample operation along the height and width dimensions of the cortical sheet. A detailed explanation of the downsampling operation and inference on such models can be found appendix A.5.

L1 unstructured pruning and downsampling were applied to progressively increase the degree of sparsity (ensuring that each degree corresponds to the same effective parameter count) and evaluate the effect on model performance. For each sparsity level, we evaluated the model’s performance (classification accuracy or perplexity) and reported the resulting performance difference from the baseline model (Figure 4).

\textbf{Estimating selectivity}

We collected layer-wise features in response to stimuli. Selectivity is then calculated using a standard method for estimating selectivity from these representations  (e.g., \cite{margalit2024unifying}):
\begin{equation}
    t = \frac{\mu_c - \mu_o}{\sqrt{\frac{\sigma_c^2}{N_c} + \frac{\sigma_o^2}{N_o}}}
\end{equation}
Where $\mu$, $\sigma$, $N$ denote the mean, standard deviation and the number of layerwise representations for the target category $c$ and other categories $o$. We used fLOC stimuli from the Grill-Spector lab to identify the category-selective regions and stimuli from the Konkle lab to obtain the topographic biases for real-world size and animacy as prior work \cite{doshi2023, margalit2024unifying}.

\textbf{Temporal window analyses:} We estimated the temporal integration window of every unit of GPT-Neo-125M using a recent developed method \cite{skrill2024large}. Briefly, this approach employs a word-swapping paradigm, measuring the difference in response magnitude for the swapped sequences. The integration window is defined as the distance-dependent change in response magnitude across multiple sequences and word swaps. For detailed methodology, we refer readers to this important study. The key equation relevant to our work is the follows. 
\begin{equation}
\theta_{\text{norm}}[\Delta] \approx c (\Delta + 1)^{-a} + (1 - c) e^{-b\Delta}
\end{equation}
$\theta_{norm}[\Delta]$ is the normalized temporal integration window and $c$ is the convex combination parameter. Intuitively $c$ is the balancing knob that controls the relative influence of two different "shapes" integration window.
$a$ represents the power-law component of the integration window. Higher values would indicate a relatively slower decline. $b$ is the exponential component of the integration window. Higher numbers would indicate a much more rapid decline. We followed exactly the same procedures outlined in the previous study to estimate these values.

\section{Results}

 \subsection{TopoNets achieve high model performance with comparable spatial topography}

How do models trained with TopoLoss, TopoNets, stack up against baseline models and other topography-inducing methods?  We first tested vision models, specifically ResNet18 trained in a supervised manner on ImageNet. This architecture allows for direct comparison with previous work \cite{margalit2024unifying, qian2024local}. (Note: data for ITN \cite{blauch2022connectivity} and All-TNNs \cite{lu2023end} are unavailable, as these architectures haven't been scaled to ImageNet). Model performance is presented against the amount of topography (smoothness, see \cite{margalit2024unifying}) in Figure 2A. We find that (1) TopoNet-ResNet18 models achieved significantly higher accuracy on ImageNet (red dots) than LLCNN (Imagenet trained supervised) and TDANN \cite{margalit2024unifying}. TDANNs were trained using a self-supervised SimCLR objective, while TopoNets were trained in a supervised manner. To ensure fairer comparison with TDANNs (see A.3 for extended discussion), we trained an additional TopoNet with topography induced in similar locations as in TDANN (in 8 locations) and compared the \textit{change} in performance from baseline (non-topographic). The drop in performance was 6\% for TDANNs versus 3\% for TopoNets. (2) TopoNets achieved comparable levels of topography (dashed gray vertical lines) as previous approaches. (3) TopoNets were trained for fewer training epochs (12\% fewer than LLCNN-G ). Even the worst performing TopoNet-ResNet18 ($\tau = 50$) was significantly better than the previous best topographic model (25\% drop for TopoNet-ResNet18-$\tau50$ compared to 41\% drop for LLCNN-G, from baseline). The model performance pareto-curve for TopoNet-ResNet18 model is shown as a black dashed line across levels of topography. Together these results indicate that models trained in a supervised manner with our new model inductive bias: TopoLoss (TopoNets) achieve substantially higher task performance than prior topographic models. TopoNets set a new standard of performance for supervised topographic ResNet18s.

 \begin{figure}
     \centering     \includegraphics[width=1\linewidth]{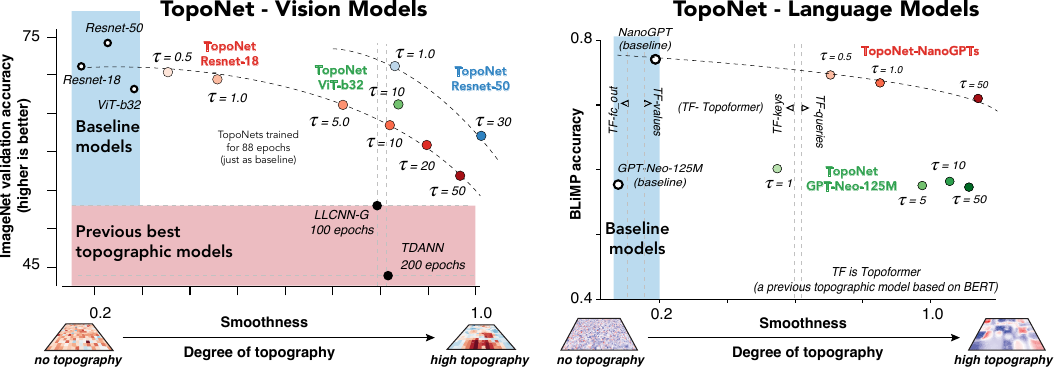}
     \caption{\textbf{TopoNets achieve higher model performance with comparable topography. A.} Estimated model topography (smoothness, x-axis) versus model performance (y-axis) for vision models (ResNet-18, ResNet-50, ViTs). The black filled dots with the dashed gray crosshairs indicate prior models. The dashed black lines indicate the pareto-curves for ResNet-18 and ResNet-50 models. \textbf{B. }Same as A. but for Language models (GPT-Neo-125M and NanoGPT). The y-axis here denotes the language model evaluation score on BLiMP. The dashed gray line indicates the reported topography from a prior study.}
     \label{fig:enter-label}
 \end{figure}

Next, we investigated whether we could develop even higher-performing TopoNets and extend the approach to transformer architectures. We trained 2 ResNet-50 models and one ViT-b32 model with TopoLoss. The model performance and topography measurements are shown for TopoNet-ResNet-50s and TopoNet-ViT-b32 as blue and green dots respectively in Figure 2. TopoNet-ResNet-50s and TopoNet-ViT both outperformed TopoNet-ResNet-18s at similar levels of topography.  Notably, our TopoNet-ResNet50-$\tau1$ achieved comparable performance as the baseline ResNet-18 model,  while exhibiting comparable levels of measured topography as prior topographic models. 

Does TopoLoss generalize to language models?  To investigate this question, we trained GPT-Neo-125M (with and without TopoLoss) on the Wikipedia dataset \cite{wikidump}. To demonstrate further scalability, we also trained NanoGPT models on 10 billion tokens from the FineWeb-Edu dataset \cite{lozhkov2024fineweb-edu}. All models were evaluated on a common evaluation measure: BLiMP \cite{warstadt2020blimp}. Our findings revealed that (1) TopoNets were comparable to the baseline (non-topographic model) for GPT-Neo-125M and were close to baseline for the scaled up NanoGPT models (Figure 2B). (2) Most TopoNets achieved higher levels of topography than Topoformers\cite{binhuraib2024topoformer} (BERT trained on IMDB) even in the layers where topography was explicitly implemented (attention Q,K,V, Figure 2B). Together these results demonstrate that TopoLoss can generalize across different model architectures (convolutional nets and transformers) and domains (vision and language). The resulting models, TopoNets, significantly outperform previous isolated efforts in vision or language alone. 

\subsection{Topography, not model performance, drives dimensionality reductions in TopoNets. Evidence across model architectures and domains}

Prior theoretical work has suggested that the brain's topography may affect non-topographic aspects of learned representations, such as the effective dimensionality \cite{durbin1990dimension, swindale1996development}. Effective dimensionality is lower when neurons are similar to each other and higher when they are independent. Studies show that 1) effective dimensionality increases with model depth and training, and 2) models with lower dimensionality better predict responses in high-level visual cortex \cite{elmoznino2024high}. However some recent work \cite{qian2024local} has raised a serious concern about this interpretation. 
Specifically it is unclear whether the reduction in dimensionality is driven by lower model performance (Hypothesis 1) or because of topography itself (Hypothesis 2). TopoNets finally allow us to test these competing hypotheses more precisely across model architectures and domains. 

We measured model dimensionality using a standard approach from previous studies \cite{del2021effective, elmoznino2024high} and examined its relationship with model accuracy (Hypothesis 1) and topography (Hypothesis 2) across both vision (ResNet-18, ResNet-50) and language models (GPT-Neo-125M, NanoGPT). These results are shown in Figure 3A. (Model performance for TDANN is included for comparison, while LLCNN is not reported as it is not yet publicly available.) We found no significant correlation between model performance and dimensionality in either vision or language models (both \textit{P} $>$  0.05, Figure 3 left column). Notably TopoNets achieved higher model performance despite equivalent levels of model dimensionality as TDANNs (Figure 3A). In contrast, dimensionality was significantly correlated with the measured topography (smoothness) in both domains (each \textit{P} $<$ 0.05). The difference between the linear relationships was statistically significant for both vision and language models (\textit{P} $<$0.05, on Fisher's z-transformed correlations). 

These results strongly support Hypothesis 2 indicating that spatial topography, rather than model performance, better accounts for reduction in the effective dimensionality of the learned representations. This analysis further highlights the broader utility of TopoNets in testing theoretical claims about the role of spatial topography across various model architectures and domains.

\begin{figure}
    \centering
    \includegraphics[width=1.\linewidth]{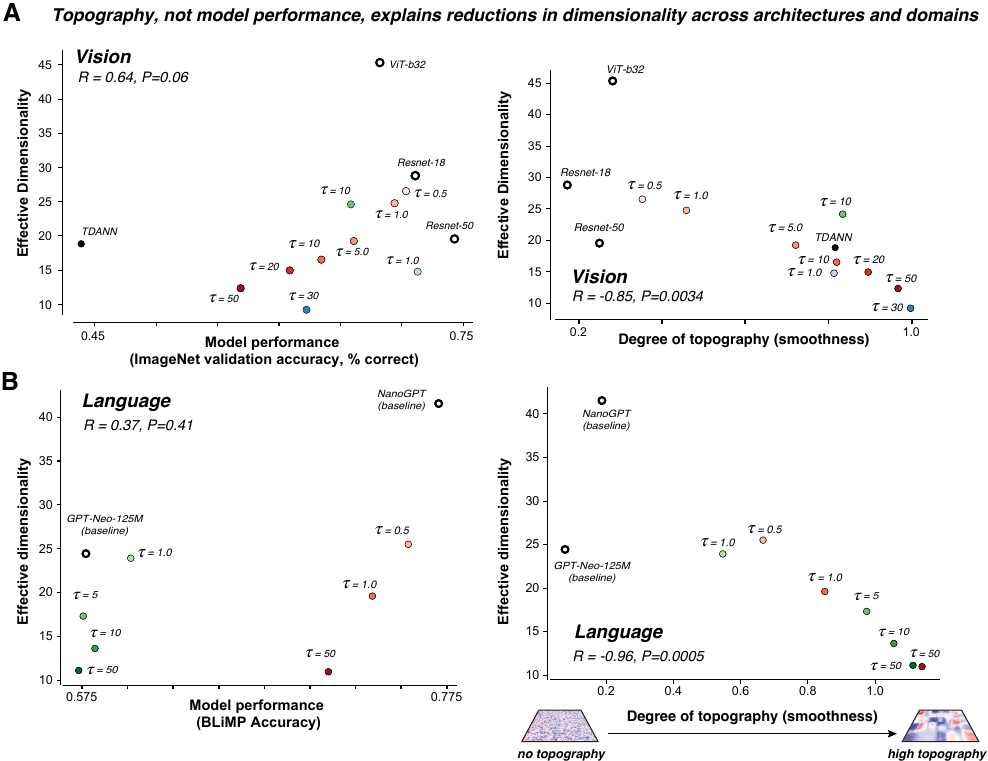}
    \caption{\textbf{Topography explains reductions in model dimensionality. A.} (Left) Model performance (ImageNet accuracy, x-axis) versus effective dimensionality for vision ResNets. (Right) Measured topography (smoothness) versus effective dimensionality for vision ResNets. \textbf{B.} Same as A, but for language transformers}
    \label{fig:enter-label}
\end{figure}

\subsection{TopoNets deliver sparse, parameter-efficient representations}
We next explored a previously unexamined application of TopoNets:  model efficiency. Brain-inspired topography encourages compact representations.  In biological systems, topographic organization results in localized and redundant information by minimizing "wiring length" (weight sparseness) and enabling more compressible "\textit{metabolically energy-efficient}" representations. Inspired by these biological principles, we asked whether TopoNets, which incorporate similar topographic constraints as brains, might exhibit two forms of efficiency: a) \textit{weight sparseness},  and b) \textit{parameter efficiency}. It is important to clarify that these measures of efficiency are distinct from model dimensionality: effective dimensionality measures the complexity of the feature representation, while weight sparseness and parameter efficiency measure the overall resource use of the model. One concerns the \textit{quality} of the learned features, the other the \textit{quantity }of the resource utilization.

We first assessed weight sparseness in TopoNets by evaluating the effect of pruning small weights using L1 unstructured pruning. Specifically, we set low-magnitude weights to zero and measured the impact of this ``lesioning" on model performance. We hypothesized that models with inherently sparser weights would be more resilient to this procedure. The results, shown in Figure 4A, illustrate the relationship between the fraction of weights lesioned (x-axis) and the corresponding drop in model performance (y-axis). As expected, as the fraction of pruned weights increased, model performance declined.  However, across both vision models (ResNet18s and ResNet50s, left and middle subplots) and language models (GPT-Neo-125Ms, right subplot), we found that TopoNets (colored dots) were more resistant to weight pruning than the baseline non-topographic models (black dots). This indicates that TopoNets produce sparser weight distributions and maintain performance more effectively when subjected to L1 unstructured pruning. 

\begin{figure}
    \centering
    \includegraphics[width=1\linewidth]{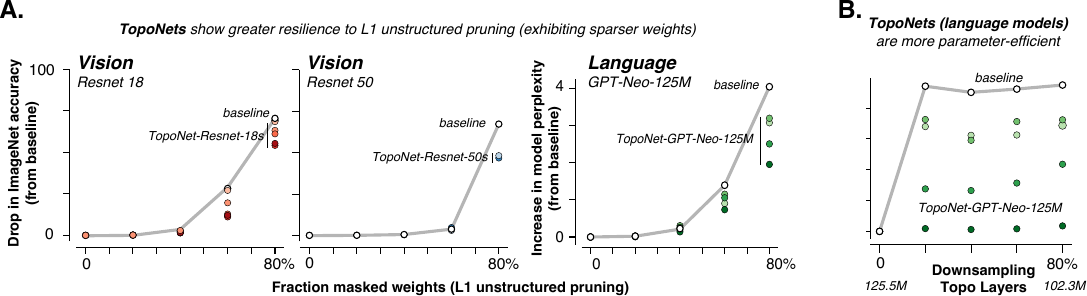}
    \caption{\textbf{Measuring the efficiency of TopoNets against baseline models: A.} Fraction of weights masked through L1 unstructured pruning (x-axis) versus the change in model performance (y-axis) for ResNet-18 (left), ResNet-50 (center), and GPT-Neo-125M (right) models. Colored circles represent TopoNets, while hollow black circles represent baseline models. The performance of the baseline models is shown by the gray line. \textbf{B.} Percentage of model weights after downsampling (x-axis) versus the drop in model performance (y-axis) for GPT-Neo-125M models.}
    \label{fig:enter-label}
\end{figure}

However, L1 pruning doesn't directly address the question of parameter efficiency. To test this aspect more directly, we downsampled the weights, thereby directly reducing the model parameter count. Due to architectural limitations, this method works only on transformer models (downsampling conv weights results in complete drop in performance to 0). These results are shown in Figure 4B. We found that TopoNets were remarkably resilient to downsampling. For instance, downsampling the weights by 80\% lowered the overall parameter count of the model from 125.6M to 102.3M parameters (a 19\% overall reduction), while maintaining performance especially at high levels of topography.  This shows that TopoNet-GPT models are significantly more parameter-efficient than baseline models. Thus TopoNets offer significant advantages in both weight sparseness and parameter efficiency. The downsampling across both GPT-Neo-125M and NanoGPT (see appendix A.10 and figure 10) results particularly suggest that TopoNets might offer a promising approach to scaling down GPT models without sacrificing task performance. This brain-inspired approach could unlock new methods for compressing large language models, providing a path for more efficient AI systems.

\subsection{TopoNets reproduce brain-like topographic signatures}

\begin{figure}
    \centering
    \includegraphics[width=1\linewidth]{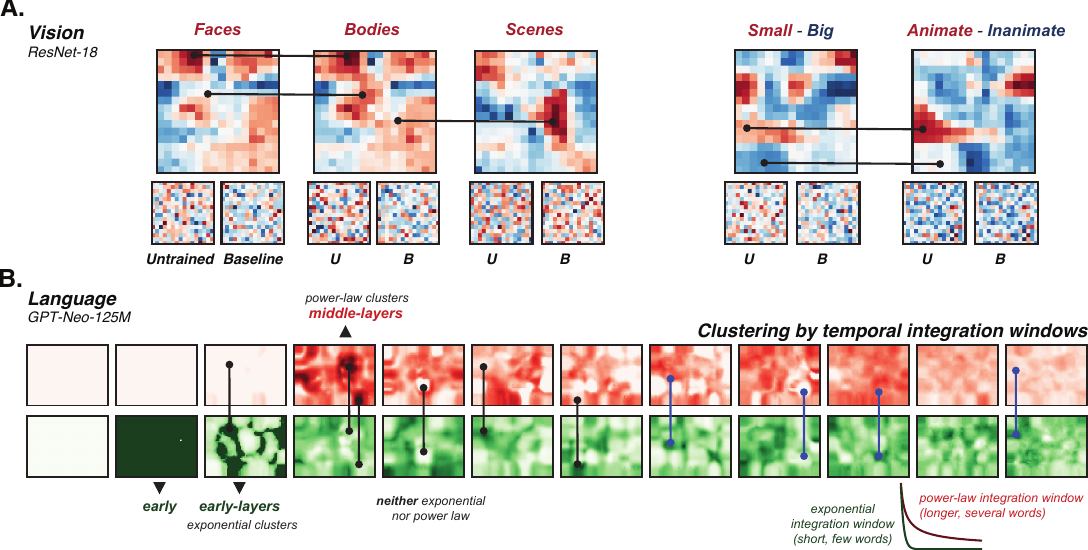}
     \caption{\textbf{TopoNets recapitulate topographic signatures observed in the visual and language cortex. A.} Topographic signatures in vision TopoNets (ResNet-18). Colormaps show t-values corresponding to selectivities for faces, bodies, scenes, real-world size, and animacy. Bold connected lines indicate the same regions across different topographic maps. Maps for the same model layer for the untrained (U) and baseline (B) models are shown below for comparison. \textbf{B.} Topographic signatures in language TopoNets. Colormaps display the strength of estimated power-law (red), exponential (green), and sentence-yoked coefficients across layers (from left to right). These coefficients indicate fast, slow and sentence-yoked temporal integration windows}
    \label{fig:enter-label}
\end{figure}

Here we evaluated the "brain-likeness" of TopoNet representations compared to other models. We evaluated vision models on 2 key neural metrics. We first tested unit-to-voxel correlations (as previously reported in \cite{margalit2024unifying}) from the Natural Scenes Dataset \cite{allen2022massive}. Model performance reached the noise ceiling for with comparable prediction accuracies to TDANN models (R = 0.54 for TDANN vs. 0.60 for TopoNet-ResNet-18 and 0.63 for TopoNet-ResNet-50, normalized to the noise ceiling across 8 subjects). Next, we compared TDANNs and TopoNets on neural metrics from BrainScore \cite{Schrimpf2020integrative,SchrimpfKubilius2018BrainScore}. TopoNets outperformed TDANN at predicting responses across all visual regions (see Table 1 for comparisons between TopoNet and TDANNs, and Appendix A.7 for all TopoNets). Taken together TopoNets  predict neural responses on a number of measures better than TDANN. We further replicated key topographic signatures in the visual cortex, such as category selectivity for faces, bodies, and scenes \cite{kanwisher2000domain, grill2004fusiform, epstein1999parahippocampal, downing2006domain, downing2001cortical}, and organizational biases for object size and animacy \cite{konkle2011canonical, konkle2013tripartite}. In Figure 5A, we show these patterns for the TopoNet-ResNet-18-$\tau10$ model. We observed that face and body selectivities were yoked together, while scene selectivity was distinct. This pattern mimics the organization observed in the FFA, FBA, and PPA in the ventral visual cortex. We also confirmed this quantitatively.  Face and scene selectivity showed a negative correlation (structural similarity = -0.41), whereas face and body selectivity were positively correlated (0.79). Additionally, TopoNets captured similar organizational biases for real-world size and animacy (structural similarity: 0.46), as seen in the brain. 



\begin{table}[!ht]
    \centering
    \small
    \begin{tabular}{l|l|l|l|l}
         & V1 & V2 & V4 & IT\\ \hline
        TopoNet-ResNet18 & {\bfseries\fontsize{12}{14}\selectfont 0.7116} & {\bfseries\fontsize{11}{11}\selectfont 0.3038} & {\bfseries\fontsize{11}{11}\selectfont 0.2923} & {\bfseries\fontsize{11}{11}\selectfont 0.5723}  \\ 
        TDANN & 0.6932 & 0.1775 & 0.2792 & 0.4259 \\ 
    \end{tabular}
    \caption{BrainScore values for TopoNet-ResNet-18 and TDANN across visual regions V1, V2, V4, and IT. The scores here averaged over several benchmarks detailed in Appendix A.7. See Appendix for all TopoNets and the baseline models.}

    \label{tab:brainscore_comparison}
\end{table}

Relatively little is known about the spatial organization of the language cortex in the brain, but some studies using fMRI \cite{lerner2011topographic, hasson2008hierarchy} and invasive recordings \cite{regev2024neural} have provided evidence for distinct temporal receptive fields. 
Based on these results, we wondered if neurons in TopoNets were clustered by their temporal integration windows. We used a new word-swapping method from a recent study \cite{skrill2024large} to investigate this in TopoNets. These temporal integration window results are  shown in Figure 5B. We replicated the expected pattern from the previous study: early layers were dominated by exponential integration dynamics, while mid-layers exhibited power-law dynamics. Interestingly, we identified three types of clusters. A) "Exponential" clusters with neurons dominated by short, exponential windows. B) "Power-law" clusters dominated by longer, power-law windows C) An intriguing cluster not explained by either exponential or power-law integration windows. These findings are illustrated in Figure 5B. Topographic maps are presented for all models (including baseline model) in Appendix Figure 8 for comparisons. To our knowledge, this is the first modeling result in topographic language models that recapitulates the experimental findings regarding clusters of temporal receptive field sizes in the language cortex in topographic-LLMs. Further work is needed to establish more precise correspondences between TopoNets and the human language system, making this an exciting direction for future research.

\section{Discussion}
Here we introduced a novel inductive bias, \textit{TopoLoss}, to train high-performing AI models that exhibit topographic organization. We created a large number of topographic AI models, TopoNets, that span both vision (ResNet-18, ResNet-50, and ViT-b32) and language (GPT-Neo-125M, NanoGPT) models. TopoNets outperformed prior topographic models on engineering benchmarks while exhibiting comparable topography (Section 3.1), addressed theoretical claims about the importance of topographic principles for low-dimensional feature representations (Section 3.2), delivered parameter-efficient representations (Section 3.3), predicted neural responses better than previous topographic models, and reproduced topographic signatures observed the brain (Section 3.4). 

Together these results tackle three questions about topography in neural networks and brains. \textit{Q1.} How do we incorporate topography in neural networks with minimal drop in model performance? TopoLoss is a new strategy for inducing topography in ANNs. It is a scalable, versatile framework, effective at inducing topography in different kinds of model architectures (conv-nets and transformers) and cognitive domains (vision and language). Importantly, TopoNets achieve minimal performance drops while successfully incorporating brain-like topography. \textit{Q2.} What is the representational consequence of brain-like topography? We demonstrate that topography, rather than task performance, shapes representations to be lower dimensional and more brain-like (Section 3.4) representations. Brain-likeness in TopoNets manifests in two key ways: (1) an improved alignment with neural data from money and human brains compared to previous topographic models (BrainScore for instance), and (2) in the ability to replicate key signatures of brain-like processing, such as category-selectivity maps in the visual cortex and temporal integration windows in the language cortex.
\textit{Q3.} Why is the brain topographic? We show that topography enhances parameter efficiency in artificial neural networks. Using complementary methods, such as lesioning (L1 pruning) and downsampling, we show that topographic networks (TopoNets) optimize computational resources. These findings highlight topography as a critical design principle for efficient and adaptive neural systems, supported by evidence from artificial neural networks as surrogate models.

\textbf{Limitations:} TopoLoss is a versatile framework compatible with foundational ANN components (linear and conv layers). But further work is required to explore the full range of its scalability particularly important in the AI setting. The model backbones in this work (specifically ResNet-18 and GPT-Neo-125M) were chosen to enable comparisons with prior research on topographic models. However, future work will need to scale these models to more complex tasks (beyond ImageNet) and larger architectures (e.g., LLaMA). That said we do not anticipate any challenges in scaling up TopoLoss to more complex architectures. Our model can be incorporated in only 2 lines of additional code (\href{https://github.com/toponets/topoloss}{\texttt{pip install topoloss}}). All preliminary tests show a mere 1-2\% performance overhead compared to baseline (non-topographic models). Another limitation is a incomplete understanding of how $\tau$, interacts with model performance and dataset complexity (though see A.4 and Figure 6). A trade-off between topography and model performance is to be expected: if $\tau$ is too high, topography may become overly rigid, limiting the model’s ability to learn useful representations. This reflects a well-known principle in computational neuroscience: a critical balance between neural constraints (topographic organization in this case) and task performance. Our framework provides an opportunity to directly test these theoretical ideas in models in future work. 

Taken together, TopoNets show that inducing topographic organization can offer competitive task performance while enhancing the efficiency and interpretability of current AI models. This work opens further interdisciplinary work in AI and neuroscience, bringing current AI systems closer to the computational strategies of the brain.

\bibliographystyle{unsrt}
\bibliography{paper}

\end{document}